\documentclass[conference]{IEEEtran}
\IEEEoverridecommandlockouts
\usepackage{cite}
\usepackage{amsmath,amssymb,amsfonts}
\usepackage{algorithmic}
\usepackage{graphicx}
\usepackage{textcomp}
\usepackage{xcolor}
\usepackage{array,balance}

\newcommand{\R}{\mathbb{R}}

\def\BibTeX{{\rm B\kern-.05em{\sc i\kern-.025em b}\kern-.08em
    T\kern-.1667em\lower.7ex\hbox{E}\kern-.125emX}}
\begin{document}

\title{Parallel Scale-wise Attention Network for Effective Scene Text Recognition\\
}

\author{\IEEEauthorblockN{Usman Sajid$^1$, Michael Chow$^2$, Jin Zhang$^2$, Taejoon Kim$^1$, Guanghui Wang$^{3}$}
\IEEEauthorblockA{$^1$\textit{Department of Electrical Engineering and Computer Science,} 
\textit{University of Kansas,
Lawrence, KS, USA, 66045} \\
$^2$\textit{Sony Interactive Entertainment Global R\&D}  \\
$^3$ \textit{Department of Computer Science},
\textit{Ryerson University,
Toronto, ON, Canada M5B 2K3}\\
Email: \{usajid, taejoonkim\}@ku.edu$^1$, \{michael.chow, jin.1.zhang\}@sony.com$^2$, wangcs@ryerson.ca$^3$}
}

\maketitle

\begin{abstract}
The paper proposes a new text recognition network for scene-text images. Many state-of-the-art methods employ the attention mechanism either in the text encoder or decoder for the text alignment. Although the encoder-based attention yields promising results, these schemes inherit noticeable limitations. They perform the feature extraction (FE) and visual attention (VA) sequentially, which bounds the attention mechanism to rely only on the FE final single-scale output. Moreover, the utilization of the attention process is limited by only applying it directly to the single scale feature-maps. To address these issues, we propose a new multi-scale and encoder-based attention network for text recognition that performs the multi-scale FE and VA in parallel. The multi-scale channels also undergo regular fusion with each other to develop the coordinated knowledge together. Quantitative evaluation and robustness analysis on the standard benchmarks demonstrate that the proposed network outperforms the state-of-the-art in most cases.
\end{abstract}


\section{Introduction}
\label{intro}
Scene text recognition aims at extracting the screen text from the given input image. It serves as a trendy task in the computer vision field. The recognition task comes up with many key challenges and issues like huge background variation in and across different images, different font styles, big fluctuation in text appearance and scale. Automated text recognition remains more desirable as manual intervention proves to be very tedious and time-consuming. Recently, deep learning-based automated methods have shown superior performance in this domain and other tasks \cite{wang2020decoupled,yu2020towards,li2019show,sajid2020zoomcount,mo2021stereo,sajid2020multi,sajid2020plug}. Some schemes perform character-level text recognition, while most methods do word/sentence level recognition. The latter one is more preferred due to its relatively easier and less tedious annotation process. 
\begin{figure}[t]
	\begin{minipage}[b]{1.0\columnwidth}
		\begin{center}
			\centerline{\includegraphics[width=1.0\columnwidth]{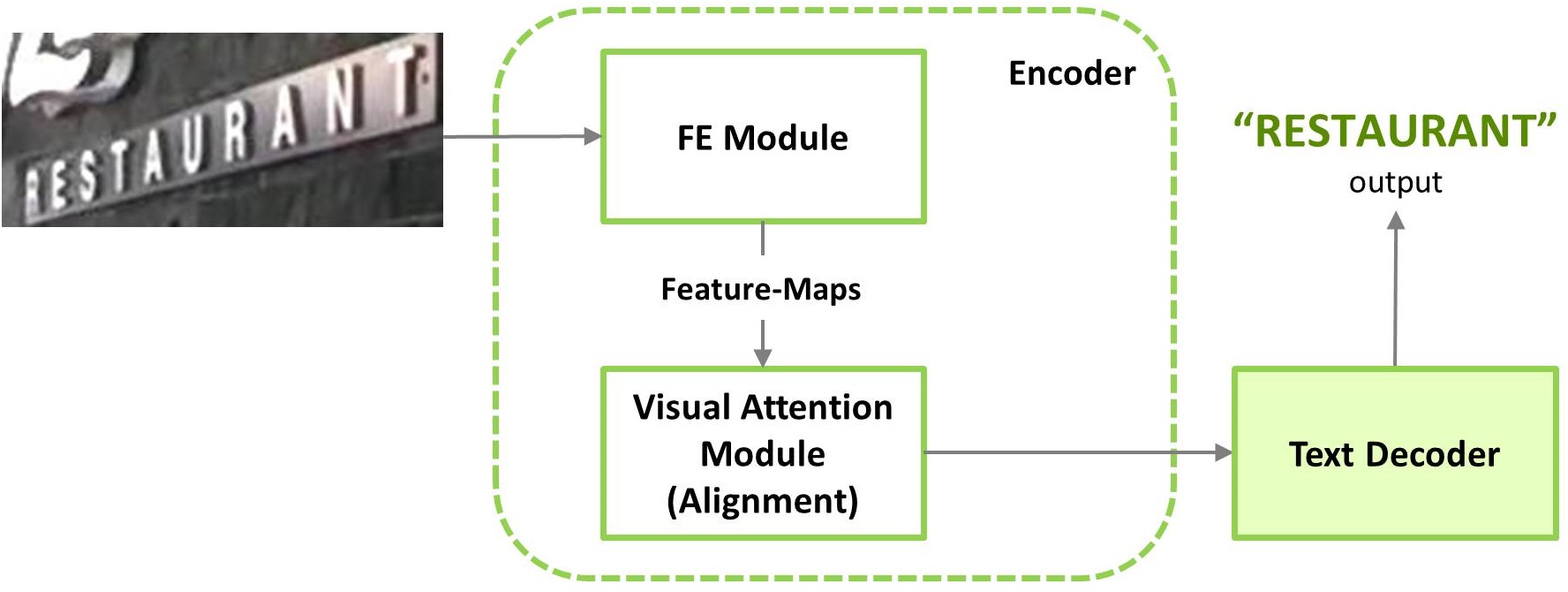}}
			\centerline{\footnotesize{(a) Sequential FE and Attention (Alignment) \cite{wang2020decoupled}}}
		\end{center}
	\end{minipage}
	\begin{minipage}[b]{1.0\columnwidth}
		\begin{center}
			\centerline{\includegraphics[width=1.0\columnwidth]{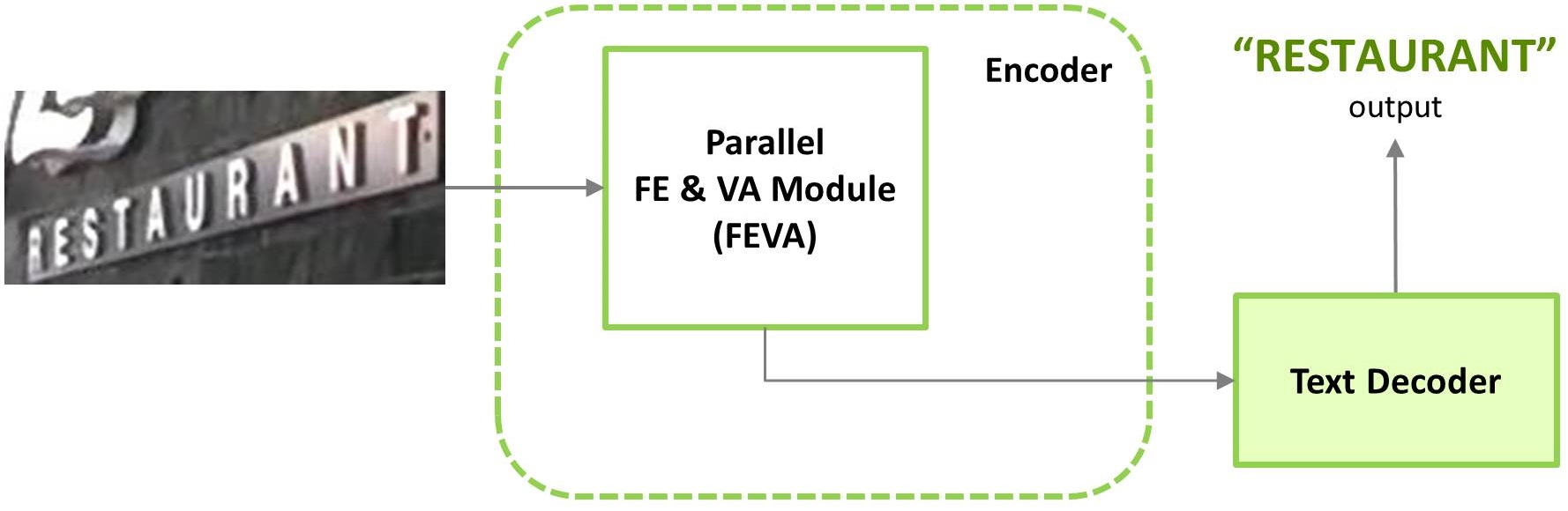}}
			\centerline{\footnotesize{(b) Parallel scale-wise FE and Alignment}}
		\end{center}
	\end{minipage}
	\vspace{0mm}
	\caption{\footnotesize{(a) The state-of-the-art Encoder-based attention mechanism \cite{wang2020decoupled} sequentially performs the feature extraction (FE) followed by the (attention) alignment module. (b) The proposed approach performs parallel FE and visual attention (VA) process on feature-maps with different scales within the encoder.}}
    \vspace{0mm}
    \label{fig:comparison}
\end{figure}

Among the best state-of-the-art deep networks, most of them \cite{wang2020decoupled,yu2020towards,cheng2017focusing,shi2016end,shi2018aster,bai2018edit,li2019show} are based on the attention mechanism \cite{bahdanau2014neural,vaswani2017attention}. The purpose of the attention mechanism is to align the text characters followed by their recognition. Generally, these methods incorporate the attention-based alignment and recognition into the decoder part of the network. However, these networks inherit an important limitation as the decoder gets highly over-burdened and sensitized with the dual task of text alignment and recognition. Consequently, it generates huge error propagation and aggregation within the decoder and thus compromises the effectiveness of the whole network. One possible solution is to decouple the attention/alignment mechanism from the decoder and integrate it with the feature extraction process inside the encoder block of the network. Recently, Wang \textit{et al.} \cite{wang2020decoupled} proposed such decoupled attention network (DAN) with promising results. However, the encoder first sequentially performs the feature extraction (FE) followed by the visual attention (VA) process as shown in Fig. \ref{fig:comparison}(a). This limits the DAN network efficacy as the attention mechanism only depends on and utilizes the final output feature-maps from the FE module. Consequently, the attention is not applied directly to each of the multi-scale feature-maps separately, but only to the final set of accumulated single-scale channels. Therefore, our focus revolves around the two main objectives in this work:

\begin{itemize}\setlength\itemsep{-0.6em}
  \item Design a scale-wise visual attention-based scene text recognition network to address the key issues and challenges in this domain.
\\  
  \item Utilize the encoder-based and scale-wise attention process in parallel to the feature extraction (FE) instead of standard sequential processing from FE to the visual attention module.
\\  
\end{itemize}

In this work, we propose a new multi-scale and scale-wise visually attended text recognition network to achieve the above objectives. As shown in Fig. \ref{fig:comparison}(b), the feature extraction and visual alignment/attention (FEVA) have been done in parallel on different scale features within a single module, followed by the recognition-focused decoder to extract the scene text. In this way, we separately attend feature-maps from different scales directly instead of just attending the final single-scale channels. Moreover, we also deploy different and simpler visual attention process in contrast to the conventional deep up- and down-scaling fully connected networks (FCN) \cite{long2015fully} based visual attention being used in DAN \cite{wang2020decoupled}. Several experiments on different standard benchmark datasets demonstrate the effectiveness of our scheme on both regular and irregular scene-texts as presented in the experiments section \ref{exps}. The main contributions of this work include:

\begin{itemize}\setlength\itemsep{-0.6em}
  \item We propose a new parallel FEVA-based encoder and multi-scale text recognition network to address the key recognition challenges and limitations in similar state-of-the-art architectures.
\\
 \item We deploy the visual attention mechanism in an effective and unique way on multiple scales to enable the network in making a clearer distinction between the foreground and background pixels. 
\\
  \item Experimental evaluation on the standard benchmark datasets demonstrates that the proposed network outperforms the state-of-the-arts in most cases on both regular and irregular scene-texts.
\\
\end{itemize}

\section{Related Work}
Text recognition problem remains a trendy topic in the computer vision field due to different challenges like varying text scale and size, partial occlusion, and non-axis aligned text. Before the deep learning era, document text recognition remained the main focus. \cite{casey1996survey} adopted the binarization process to extract the segmented text characters. But these methods are not applicable to scene-text due to different nature of issues like varying scale and style, and complex background. Most of the classical recognizers utilized the low-level information including the connected components \cite{neumann2012real}, gradients descriptors (HoG) based on some feature-extraction mechanism \cite{wang2011end}. Recently, deep-learning based method hugely surpass and outperform the traditional methods. They are categorized as segmentation-relying and segmentation-less text recognizers.

Segmentation-based methods undergo character-wise detection followed by the word formation. \cite{bissacco2013photoocr} designed five hidden fully-connected layers and ReLU Units \cite{nair2010rectified} with softmax-based classification. \cite{wang2012end} developed a convolutional neural network (CNN) with convolution and average pooling layers and the non-maximum suppression for character-wise text recognition. \cite{jaderberg2014synthetic} used weight-shared CNN for three sub-tasks of dictionary, character sequence, and bag-of-N-gram encoding to perform the text recognition.

The segmentation-less schemes directly recognize the whole word or sentence from the given input image. \cite{jaderberg2016reading} performed a CNN-based 90,000-way classification, where each category/class corresponds to one whole word. Shi \textit{et al.} \cite{shi2016end} integrated the convolutional neural network (CNN) and recurrent neural network (RNN)-based scheme to obtain the string features, and the Connectionist Temporal Classification (CTC)-based decoder to finally yield the recognized text. \cite{shi2016robust} employed the attention mechanism for text alignment before the recognition. Most following methods utilize the attention mechanism \cite{bahdanau2014neural,vaswani2017attention} in one way or the other. Cheng \textit{et al.} \cite{cheng2017focusing} designed the deep focused attention network (FAN) after observing and aiming to address the ``attention-drift" problem in the recognition process, but it requires character-level annotations. \cite{shi2016robust,luo2019moran,zhan2019esir} aimed at addressing non-axis aligned and distorted text via an attention-based mechanism. \cite{su2014accurate,su2017accurate} utilized the RNNs with Long Short Term Memory (LSTM) networks to perform the sequential word recognition. \cite{he2016reading} integrated the CNN and RNN to design the deep-text recurrent network (DTRN) for recognizing text. Shi \textit{et al.} \cite{shi2018aster} explicitly handled the text rectification by using the control points based rectification module and also applied the attention-based bi-directional LSTM decoder for text prediction. Li \textit{et al.} \cite{li2019show} proposed a simple LSTM-based encoder-decoder framework via the 2D attention process. Wang \textit{et al.} \cite{wang2020decoupled} designed the decoupled attention network (DAN) that performed the text alignment 
via convolution-based visual attention. Yu \textit{et al.} \cite{yu2020towards} proposed the semantic reasoning network (SRN) for irregular scene-text that fuses the visual attention and semantic context modules while avoiding the RNN-based sequential processing.

Although these schemes produce good results, yet they fail to utilize the promising and beneficial attention mechanism explicitly on different multi-scale features. In this work, we work towards utilizing the multi-scale feature-extraction and visual-attention in parallel for better efficacy.

\begin{figure*}
	\begin{minipage}[b]{1.0\textwidth}
		\begin{center}
			\centerline{\includegraphics[width=0.95\textwidth]{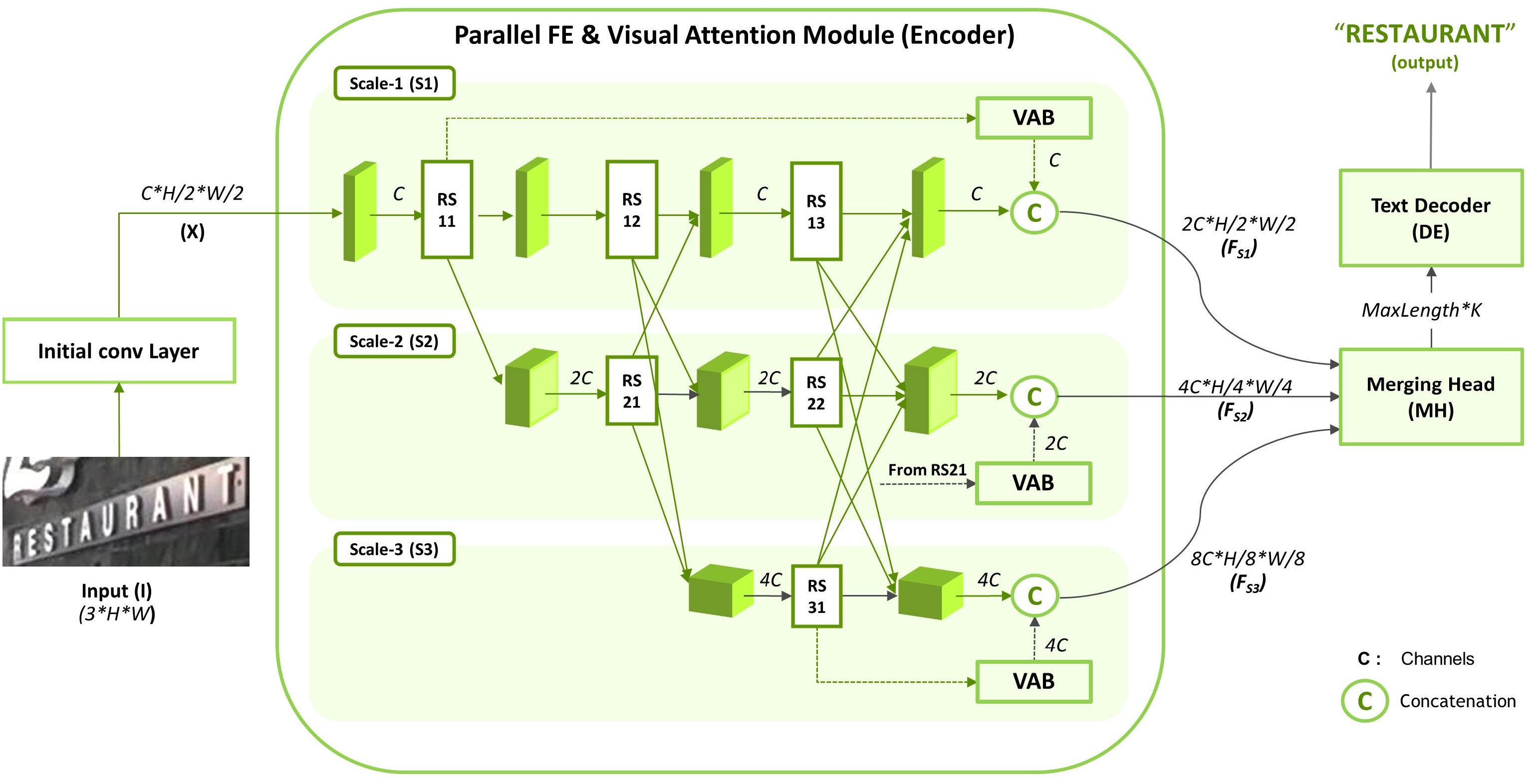}}
		\end{center}
	\end{minipage}		
    \vspace{-6mm}
	\caption{The proposed text recognition network. Initially extracted features ($X$) from the input image ($I$) first pass through the multi-scale feature-extraction (FE) and visual attention (VA) based encoder (EN). The encoder performs both intra-scale processing and inter-scale fusion between three scale-modules (S1, S2, S3). During the intra-scale processing, the channels are processed with the residual connections-based residual structures (RS) as well as undergo the VA mechanism via the visual attention block (VAB), followed by their concatenation to produce the respective scale-module output. Consequently, the encoder outputs three sets of feature-maps that go through the merging head (MH) for channel and resolution adjustment. Finally, the text decoder (DE) outputs the recognized text character-wise.}
    \vspace{-3mm}
    \label{fig:fig2_architecture}
\end{figure*}

\section{Proposed Approach}
The paper proposes a new scene-text recognition network to address the major recognition challenges as detailed in Sec. \ref{intro}, as well as performs the visual attention explicitly on multi-scale feature-maps in parallel. The proposed network, as shown in Fig. \ref{fig:fig2_architecture}, downscales the input image I ($\in \R^{3*H*W}$) resolution by half ($C*\frac{H}{2}*\frac{W}{2}$) using the initial convolutional layer. Here ($C=Channels, H=Height, W=Width$). The resultant feature-maps go through the text encoder (EN). The EN block comprises three parallel multi-scale modules (S1, S2, S3) with each module handling one specific scale. Each multi-scale module also visually attends the feature-maps in parallel to their conventional deep layers-based processing, followed by the concatenation together to generate their respective output. The visual attention helps the model to have a clearer understanding of the foreground and background pixels. Inspired by the high-resolution networks \cite{wang2020deep,sun2019deep}, these multi-scale modules also fuse their channels at regular intervals to develop the accumulated knowledge together. The encoder outputs three multi-scale channels ($F_{S1},F_{S2},F_{S3}$) that are merged together via the merging head (MH). The text decoder (DE) finally outputs the recognized text. The proposed network architecture consists of three major components: Text Encoder (EN), Merging Head (MH), and Text Decoder (DE) as detailed next.

\subsection{Encoder (EN)}
The purpose of the encoder is to simultaneously perform the feature extraction (FE) and visual attention/alignment (VA) on the multi-scale feature-maps. The input channels ($X\in \R^{C*\frac{H}{2}*\frac{W}{2}}$) pass through three multi-scale modules(S1, S2, S3) to finally yield three respective output feature-maps with different dimensions. The encoder processes the input feature-maps as follows:

\vspace{0mm}
\begin{equation}
\label{eq1}
(F_{S1},F_{S2},F_{S3})=Encoder(X),
\end{equation}
where $F_{S1}\in \R^{2C*\frac{H}{2}*\frac{W}{2}}$, $F_{S2}\in \R^{4C*\frac{H}{4}*\frac{W}{4}}$, $F_{S3}\in \R^{8C*\frac{H}{8}*\frac{W}{8}}$ and \textit{C} indicates the total number of input channels. As we move from S1 to S3, the number of channels becomes twice as many as their subsequent upper scale. Similarly, the feature-map resolution (scale) decreases to half with each scale-module as we move from S1 to S3. It may be noted that each scale module keeps the channel resolution the same throughout that module \cite{sun2019deep,wang2020deep}.

\subsubsection{Intra-Scale Processing}
Within every scale module (S1, S2, S3), the input channels pass through one or more residual structures (RS) and the visual attention process. 

\textbf{Residual Structure (RS).} Each RS block comprises of five residual units (RU). The RU unit is a 3-layered residual building block as given in \cite{he2016deep} that contains three convolution layers ($1\times1,3\times3,1\times1$) and a residual connection. After every convolution operation in the paper, we deploy the Batch-Normalization (BN) \cite{ioffe2015batch} and the ReLU activation \cite{nair2010rectified} unless stated otherwise. The RS blocks are denoted as \textit{RS(xy)}, where \textit{x} denotes the scale-module number (1,2 or 3) and \textit{y} indicates their location or index within that module (starting from left to right). Thus, \textit{RS12} denotes the second RS block in the S1 scale-module.

\begin{figure}[t]
	\begin{minipage}[b]{1.0\columnwidth}
		\begin{center}
			\centerline{\includegraphics[width=1.0\columnwidth]{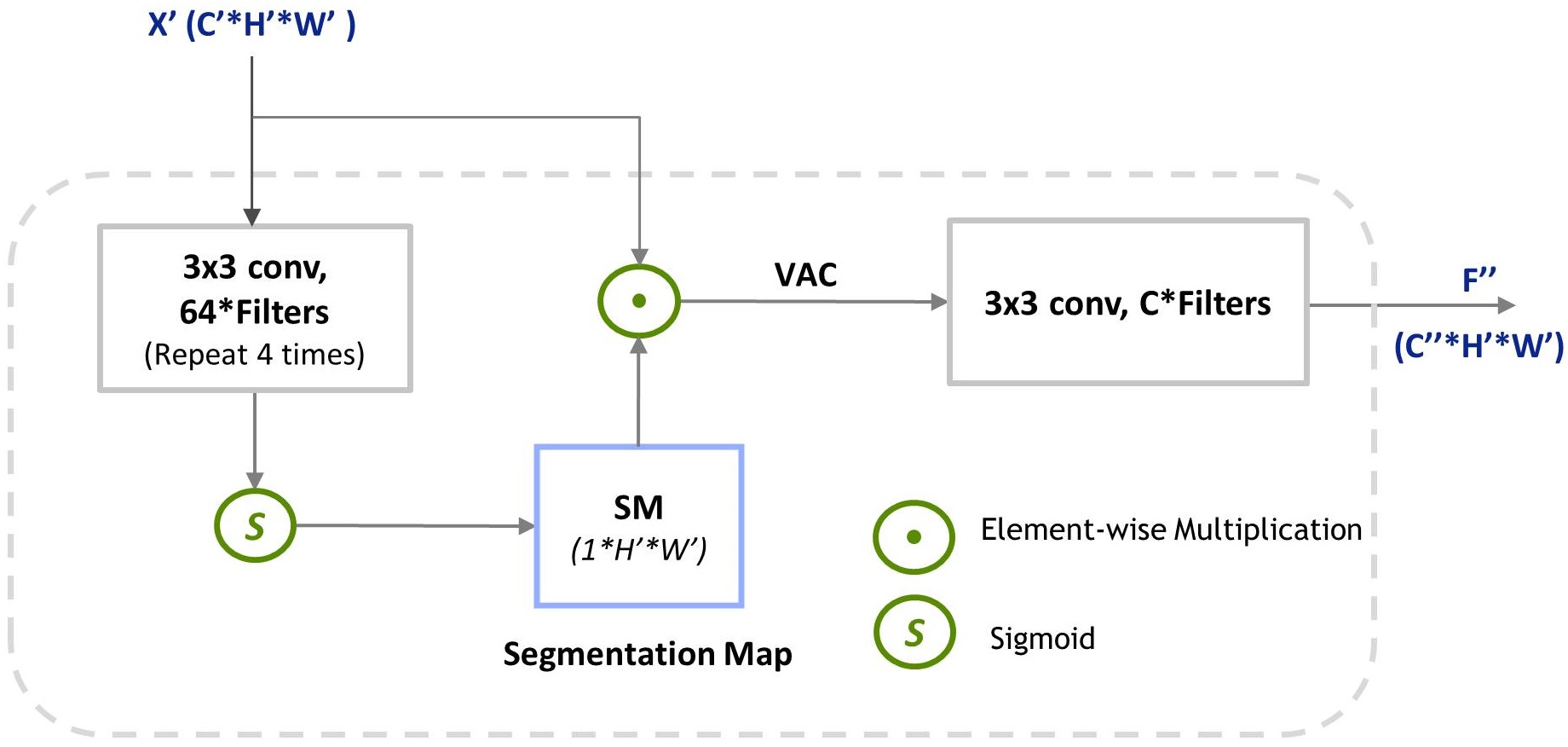}}
		\end{center}
	\end{minipage}
	\vspace{-7mm}
	\caption{\footnotesize{Visual Attention Block (VAB). The input channels go through the convolution operation four repeated times. Subsequent single-filter $1\times1$ convolution and sigmoid function give the segmentation map (SM). The SM undergoes the element-wise multiplication with the original input channels to yield the visually attended channels (VAC) that are channel-adjusted to become the VAB output.}}
    \vspace{5mm}
    \label{fig:attntn_module}
\end{figure}

\textbf{Visual Attention Block (VAB).} The scale-modules (S1, S2, S3) also visually attend (align) their feature-maps independently. This helps the network in making a better understanding regarding the foreground and background image pixels at different feature-scales. The first RS block output channels ($\in \R^{C'*H'*W'}$) in any scale-module undergo the attention mechanism via the \textit{VAB} block. The attended feature-maps are then concatenated back at the end of the respective scale-module. The VAB process is shown in Fig. \ref{fig:attntn_module}, where the input feature-maps first go through the five consecutive convolution layers. Next, a single feature-map is obtained via a simple $1\times1$ convolution operation. The sigmoid function is then applied on the resultant channel to obtain the segmentation map ($SM\in \R^{1*H'*W'}$). The SM undergoes element-wise multiplication with the original input feature-maps to yield the visually-attended channels (VAC). The VAC feature-maps serve as the VAC module final output after being channel-adjusted via the $3\times3$ convolution operation. The VAB input feature-maps $X'\in \R^{C'*H'*W'}$ get visual attention as follows:

\vspace{0mm}
\begin{equation}
\label{eq2}
F''=VAB(X'),
\end{equation}
where $F''\in \R^{C''*H'*W'}$ and we set $C''=C'$. This attention process is different from the conventional and complex convolutional and deconvolutional layers based mechanism \cite{wang2020decoupled}, and proves to be more effective as demonstrated in the experiments Sec. \ref{exps}.

\begin{figure}
	\begin{minipage}[b]{1.0\columnwidth}
		\begin{center}
			\centerline{\includegraphics[width=1.0\columnwidth]{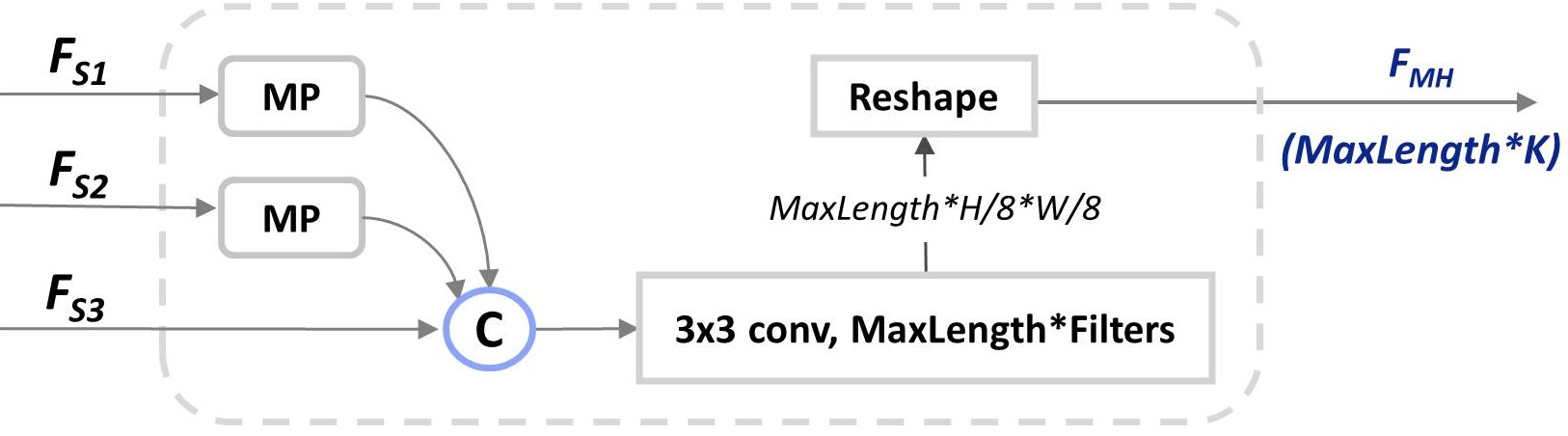}}
		\end{center}
	\end{minipage}
	\vspace{-7mm}
	\caption{\footnotesize{Merging Head (MH). The higher scale-modules (S1,S2) output
channels ($F_{S1}$, $F_{S2}$) are max-pooled before concatenation with the S3 scale-module output. Convolution and reshaping operations finally output the K-dimensional vectors with $MaxLength$ such vectors in total.}}
    \vspace{0mm}
    \label{fig:mergehead}
\end{figure}

\subsubsection{Repetitive Inter-scale Fusion}
Inspired by the high-resolution networks \cite{wang2020deep,sun2019deep}, the scale-modules (S1, S2, S3) also fuse channels with each other on regular intervals. It enables the network to form the accumulated and coordinated knowledge from the multi-scale channels and learn the valuable information better. To fuse the higher-scale source channels into the lower-scale target feature-maps, they undergo the ($n+1$) times $3\times3$ convolution operation (with stride: 2, padding: 1). Here $n$ ($=0,1$) denotes the number of scale-modules in-between the source and target scale-modules. Thus, fusion from S1 channels into S3 requires two such convolution operations on S1 scale feature-maps to down-scale them to the S3 scale. Similarly, the lower-to-higher scale fusion requires the bilinear upsampling of the lower-scale source feature-maps. No re-scaling transformation is done when the source and target scale-modules are the same. Once all source channels have been adjusted for channel quantity and target scale, they undergo the summation-based fusion with the target channels to obtain the fused feature-maps. 

\subsection{Merging Head (MH)}
The encoder outputs three separate sets of feature-maps ($F_{S1},F_{S2},F_{S3}$) from the respective scale-modules (S1,S2,S3). The merging head (MH) combines them to output the feature-maps to be used for the text decoding. The MH block, as shown in  Fig. \ref{fig:mergehead}, down-samples the S1 and S2 output channels using the max-pooling (MP) operation, so as to rescale them to the S3 output channels ($F_{S3}$) resolution. Next, they are concatenated together followed by the channel-adjustment via the convolutional layer. The resultant channels ($\in \R^{MaxLength*\frac{H}{8}*\frac{W}{8}}$) are reshaped into the K-dimensional vectors to give the MH final output ($F_{MH}\in \R^{MaxLength*K}$). Thus, the input channels are merged as follows:

\vspace{0mm}
\begin{equation}
\label{eq3}
F_{MH}=MH(F_{S1},F_{S2},F_{S3}),
\end{equation}

Here, the MaxLength refers to the maximum length of text characters to be recognized. The output vectors are then routed to the text decoder for further processing.

\begin{figure}[t]
	\begin{minipage}[b]{1.0\columnwidth}
		\begin{center}
			\centerline{\includegraphics[width=1.0\columnwidth]{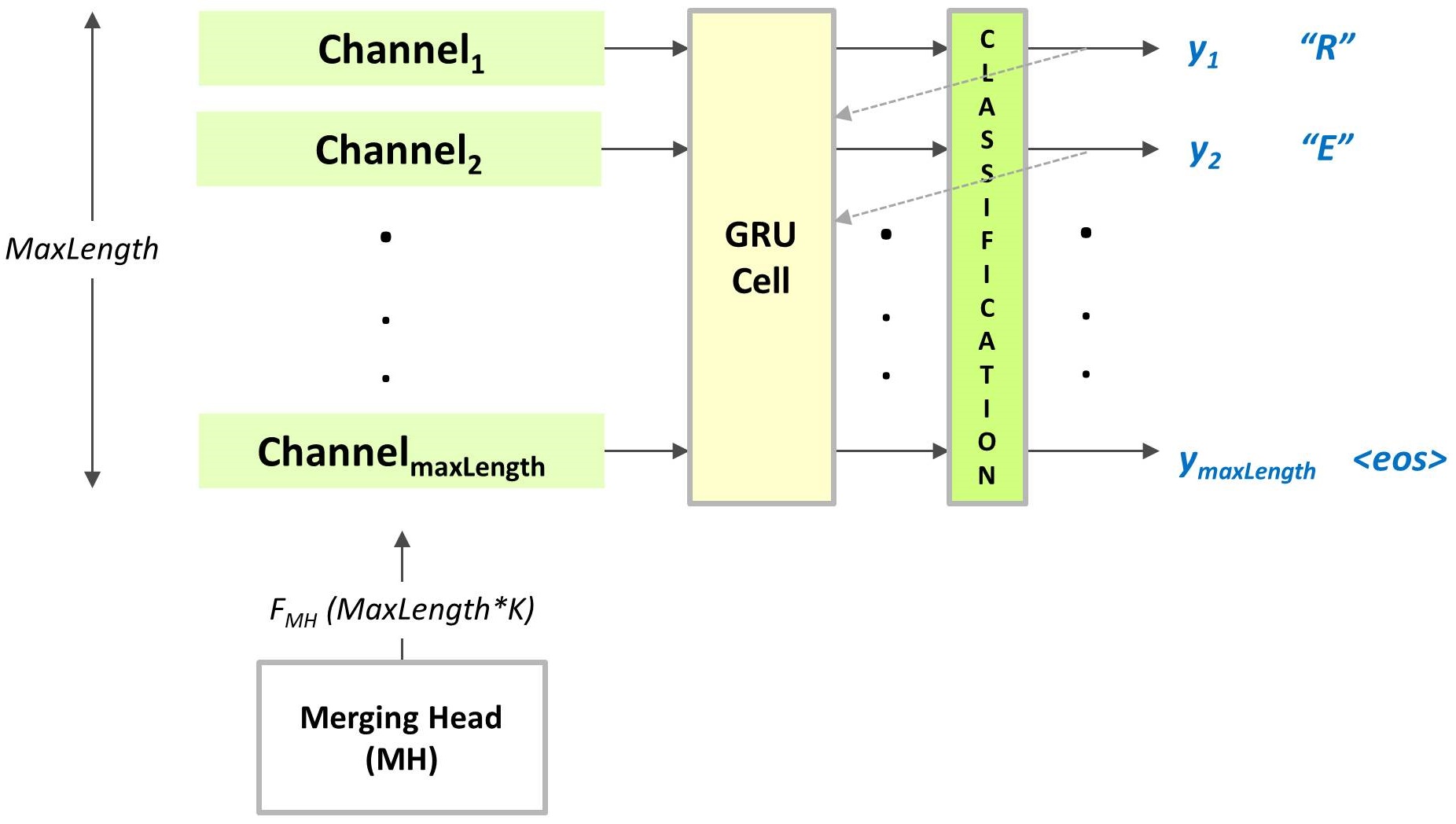}}
		\end{center}
	\end{minipage}
	\vspace{-7mm}
	\caption{\footnotesize{Text Decoder (DE). The decoder predicts the recognized text character-by-character via the GRU and the classification layers. Here \textit{eos} means the end-of-sequence character.}}
    \vspace{0mm}
    \label{fig:decoder}
\end{figure}

\setlength{\tabcolsep}{2.0pt}
\begin{table*}[t]\small

 \vspace{0mm}
	\caption{\footnotesize Quantitative Evaluation on the Standard Benchmarks. The results demonstrate that the proposed scheme is the most effective in most cases as compared to the SOTA methods on the recognition accuracy. The bold and underlined numbers indicate the best and the second-best methods respectively.}

	\begin{center}
	\begin{tabular}{|c|c|c|c|c|c|c|c|c|}
    \hline

& Rect. & \multicolumn{4}{c|}{Regular Datasets} & \multicolumn{3}{c|}{Irregular Datasets}\\ \hline
Method &  & IIIT-5K & SVT & IC03  & IC13 & IC15  & SVT-P & CUTE80\\ \hline

Jaderberg \textit{et. al}\cite{jaderberg2016reading} &   & - & 80.7 & 93.3  & 90.8 & -   & - & -  \\ 
Jaderberg \textit{et. al}\cite{jaderberg2014deep} &   & -  & 71.7 & 89.6  & 81.8 & -   & - &  \\ 
Shi \textit{et. al}\cite{shi2016end} & & 81.2 & 82.7  &  91.9 & 89.6 & - & - & -  \\ 
Lyu \textit{et. al}\cite{lyu20192d} &  & 94.0 & 90.1 &  94.3 & 92.7 & 76.3   & 82.3  & 86.8 \\ 
Xie \textit{et. al}\cite{xie2019aggregation} & & 82.3 & 82.6 & 92.1 & 89.7  & 68.9   & 70.1 & 82.6 \\ 
Liao \textit{et. al}\cite{liao2019scene}   & & 91.9 & 86.4 &  -  & 91.5 & -   & - & - \\ 

Cheng \textit{et. al}\cite{cheng2017focusing}   & & 87.4  & 85.9 &  94.2  & 93.3 & 70.6   & - & - \\ 

Cheng \textit{et. al}\cite{cheng2018aon} & & 87.0  & 82.8  & 91.5 & - & 68.2 & 73.0 & 76.8    \\ 

Bai \textit{et. al}\cite{bai2018edit} & & 88.3  & 87.5 &  \underline{94.6} & 94.4 & 73.9   & - & - \\ 
Yang \textit{et. al}\cite{yang2017learning} & & -  & - &  - & -  & -   & 75.8  & 69.3 \\ 
Shi \textit{et. al}\cite{shi2018aster} & \checkmark & 93.4   & 89.5 & 94.5 & 91.8 & 76.1   & 78.5 & 79.5  \\ 

Zhan \textit{et. al}\cite{zhan2019esir} & \checkmark &   &  & - & 91.3 & 76.9   & 79.6 & 83.3 \\ 

Yang \textit{et. al}\cite{yang2019symmetry} & & 93.3 & 90.2  &  91.2 & 93.9 & 78.7 & 80.8 & \underline{87.5} \\ 

Li \textit{et. al}\cite{li2019show} & & 91.5 & 84.5 &  - & 91.0 & 69.2 & 76.4 & 83.3 \\ 
Liao \textit{et. al}\cite{liao2019mask} &  & 93.9 & 90.6 &  - & 95.3  & 77.3 & 82.2 & \textbf{87.8}\\ 

Wang \textit{et. al}\cite{wang2020decoupled} & & 94.3  & 89.2 &  \textbf{95.0} & 93.9  & 74.5 & 80.0 & 84.4 \\ 
Yu \textit{et. al}\cite{yu2020towards} & & \underline{94.8} & \textbf{91.5} &  - & \underline{95.5}  & \underline{82.7} & \underline{85.1} & \textbf{87.8} \\ \hline \hline

\textbf{Ours} &  & \textbf{95.9}  & \underline{90.8} &  \underline{94.6}  &  \textbf{96.3} & \textbf{83.9}  & \textbf{86.0} & 86.9  \\ \hline


	\end{tabular}
	\end{center}
	   
	\label{table:quant_results}
    \vspace{-5mm}
\end{table*}

\subsection{Text Decoder (DE)}
The responsibility of our text decoder is to perform recognition only. That makes it more focused on one task rather than the dual task of text alignment and recognition. We adopted the text decoder from the DAN  network \cite{wang2020decoupled}. As shown in Fig. \ref{fig:decoder}, the MH output channels ($F_{MH}\in \R^{MaxLength*K}$) go through the GRU \cite{cho2014learning} cell one-by-one at time ($t'=1,2,3,...MaxLength$) as K-dimensional vectors. The classification layer outputs the recognized text character at time $t'$ with the output $p(y_{t'})$ as follows:

\vspace{0mm}
\begin{equation}
\label{eq4}
p(y_{t'})=\mathrm{softmax}(w*hidden_{t'}+b),
\end{equation}
where $hidden_{t'}$ denotes the GRU cell hidden state, given as follows:

\begin{equation}
\label{eq5}
hidden_{t'}=GRU((embd_{t'-1},Channel_{t'}),hidden_{t'-1}),
\end{equation}
where $embd_{t'-1}$ is the embedding belonging to the previous classification $y_{t'-1}$. The network loss function is defined as follows:

\begin{equation}
\label{eq6}
L=- \sum_{t'=1}^{T'}\log P(y'_{t'}|Input,\theta)
\end{equation}
where $P$ indicates the prediction probability, $y'_{t'}$ is the actual or ground-truth text character at time $t'$ and $\theta$ denotes the learnable parameters of the network.

\section{Quantitative and Qualitative Evaluation}
\label{exps}
This section deals with the experimental analysis and comparison of the proposed network. First, we discuss the quantitative evaluation on seven standard benchmark datasets followed by the ablation study. We conclude with the visual analysis.

\subsection{Experiments on Standard Benchmarks}

\noindent \textbf{Datasets.} 
To evaluate the efficacy of the proposed network, we test on seven different scene-text datasets. They are either regular (IIIT-5k \cite{mishra2012scene}, IC03 \cite{lucas2005icdar}, IC13 \cite{karatzas2013icdar}, SVT \cite{wang2011end}) or irregular (IC15 \cite{karatzas2015icdar}, SVT-P \cite{quy2013recognizing}, CUTE80 \cite{risnumawan2014robust}) scene-text datasets.

\textbf{IIIT-5k} \cite{mishra2012scene} is an internet-based scene-text dataset that contains 3,000 cropped text images for testing.

\textbf{Street View Text (SVT)} \cite{wang2011end} comprises of 647 text-based test images collected via Google Street View. For diversity and variation, drastic corruption has been incorporated in the form of noise, blurriness, and low resolution.

\textbf{ICDAR 2003 (IC03)} \cite{lucas2005icdar} has 251 scene-text images with 867 test bounding boxes. As per the standard protocol \cite{wang2011end}, 860 cropped images have been retained after removing words with non-alphanumeric or less than 3 characters.

\textbf{ICDAR 2013 (IC13)} \cite{karatzas2013icdar} is a regular scene-text dataset that contains total 1,015 cropped images, and most of them come from the IC03 dataset. Using the standard practice as given in \cite{wang2011end}, images with non-alphanumeric or less than three characters have been filtered out.

\textbf{ICDAR 2015 (IC15)} \cite{karatzas2015icdar} contains irregular scene-text images taken via the Google Glasses with slight focusing and positioning. Only 1,811 test images have been utilized after removing some with extreme distortions as part of the standard pre-processing practice \cite{cheng2017focusing}.

\textbf{SVT-P} \cite{quy2013recognizing} is an irregular scene-text dataset with 639 cropped images taken from Google street view. Mostly, they are single-angle based and highly perspective-distorted images.

\textbf{CUTE80} \cite{risnumawan2014robust} mainly deals with curved scene-text and consists of 80 images. We cropped 288 test samples from these high-resolution images using their bounding-box annotations.

\noindent \textbf{Implementation Details.} The input image gets resized with fixed height of 32 pixels and width up to 128 based on the aspect ratio. The proposed network is trained using two synthetic datasets until convergence: Synth90k \cite{jaderberg2014synthetic} and SynthText \cite{gupta2016synthetic}. A batch size of 64 has been used with 32 images each from Synth90k and SynthText. The value of total channels (C) in the encoder has been set to 32, so the scale-modules (S1, S2, S3) contain (32, 64, 128) channels respectively after every intra-scale processing step. MaxLength is set to 25, and the total number of character classes is 94 including the upper- and lower-case alphabets, 0-9 digits, and 32 ASCII punctuation symbols. The total decoder hidden units are set to 256. The ADADELTA-based optimization \cite{zeiler2012adadelta} has been employed with the initial learning rate value of 1.0 and decreased to 0.1 from the fourth epoch.

\setlength{\tabcolsep}{2.0pt}
\begin{table}[t]\small
\vspace{-2mm}
\caption{\footnotesize Ablation studies on the proposed network. Several experiments on different components of the proposed network indicate their vitality.}

	\begin{center}
	\begin{tabular}{|c|c|c|c|c|c|c|}
    \hline
    
 \multicolumn{7}{|c|}{VAB Block Effect}\\ \hline
 
& IIIT5k & SVT   & IC13 & IC15 & SVT-P  & CUTE80 \\ \hline
 w/o VAB & 86.6 & 81.9 & 88.8 & 77.4 & 80.7 & 75.2  \\ 
 \textbf{w VAB (ours)} & \textbf{95.9} & \textbf{90.8} & \textbf{96.3} & \textbf{83.9}  & \textbf{86.0} & \textbf{86.9}    \\ \hline \hline
 
 \multicolumn{7}{|c|}{Number of Residual Units (RUs) per RS Block}\\ \hline
    & IIIT5k & SVT & IC13 & IC15 & SVT-P  & CUTE80 \\ \hline
   1 & 61.5 & 58.7 & 61.2 & 55.3 & 59.5  & 62.1  \\ 
   2 & 72.6 & 65.5 & 68.0 & 61.0 & 65.9 & 67.0  \\ 
   3 & 83.0 & 76.1 & 77.9 & 71.7 & 74.8 & 77.5  \\ 
   4 & 90.1 & 84.9 & 83.5 & 79.6 & 82.3 & 84.2  \\ 
 \textbf{5 (ours)} & \textbf{95.9} & \textbf{90.8} & \textbf{96.3} & 83.9  & \textbf{86.0} & \textbf{86.9} \\ 
 6 & 94.3 & 88.8 & 95.6 & \textbf{84.0} & 85.5 & 86.2
 \\ \hline \hline

\multicolumn{7}{|c|}{S2 and S3 scale-modules Effect}\\ \hline
Scale-Modules & IIIT5k & SVT & IC13 & IC15 & SVT-P  & CUTE80 \\ \hline
   S1 only & 87.4 & 81.5 & 90.1 & 79.9 & 81.3 & 82.1  \\ 
   S1,S2 only & 92.9 & 87.3 & 93.6 & 82.2 & 83.7 &  84.5 \\ 
\textbf{S1,S2,S3 (ours)} & \textbf{95.9} & \textbf{90.8} & \textbf{96.3} & \textbf{83.9}  & 86.0 & \textbf{86.9}\\ 
   S1,S2,S3,S4 & 95.1 & 90.6 & 85.4 & 82.9 & \textbf{86.5} & 86.0   
 \\ \hline \hline
 
\multicolumn{7}{|c|}{MaxLength Effect}\\ \hline
MaxLength & IIIT5k & SVT & IC13 & IC15 & SVT-P  & CUTE80 \\ \hline
  \textbf{25 (ours)} & \textbf{95.9} & \textbf{90.8} & \textbf{96.3} & \textbf{83.9}  & \textbf{86.0} & \textbf{86.9} \\ 
   50 & 95.5 & 90.7 & 96.2 & 83.7 & 85.9  & 86.9  \\ 
   75 & 95.6 & 90.7 & 96.1 & 83.8 & 86.0 &  86.8 \\ 
   100 & 95.8 & 90.6 & 96.2 & 83.6 & 85.9 & 86.9
 \\ \hline 
 
	\end{tabular}
	\end{center}
	\label{table:abl_results}
	\vspace{-4mm}
\end{table}

\noindent \textbf{Experimental Evaluation.} 
Here, we compare our method quantitatively with the recent best networks. The comparison is done without using the lexicon information as it is generally the case in practice. As per the standard convention, the evaluation is done using the case-insensitivity for word accuracy computation. The results are shown in Table \ref{table:quant_results}, where our method outperforms other methods on 4 out of 7 datasets while performing reasonably competitive on the remaining three benchmarks. In comparison to the specifically designed rectification-based methods \cite{shi2018aster,zhan2019esir,luo2019moran}, our model gives better or competitive results without any rectification. For the regular scene-text dataset (IIIT-5K and IC13), we obtain an increase of ($0.8\%$ and $1.1\%$) respectively. While for the irregular scene-text datasets (IC15 and SVT-P), the proposed network improves the accuracy by ($1.4\%$ and $1.0\%$) respectively. The accuracy boost is mainly due to the inclusion of multi-scale visual attention and inter-scale fusion within the encoder. It is empirically shown during the ablation study as given in following paragraphs.

\begin{figure}[t]
	\begin{minipage}[b]{1.0\columnwidth}
		\begin{center}
			\centerline{\includegraphics[width=1.0\columnwidth]{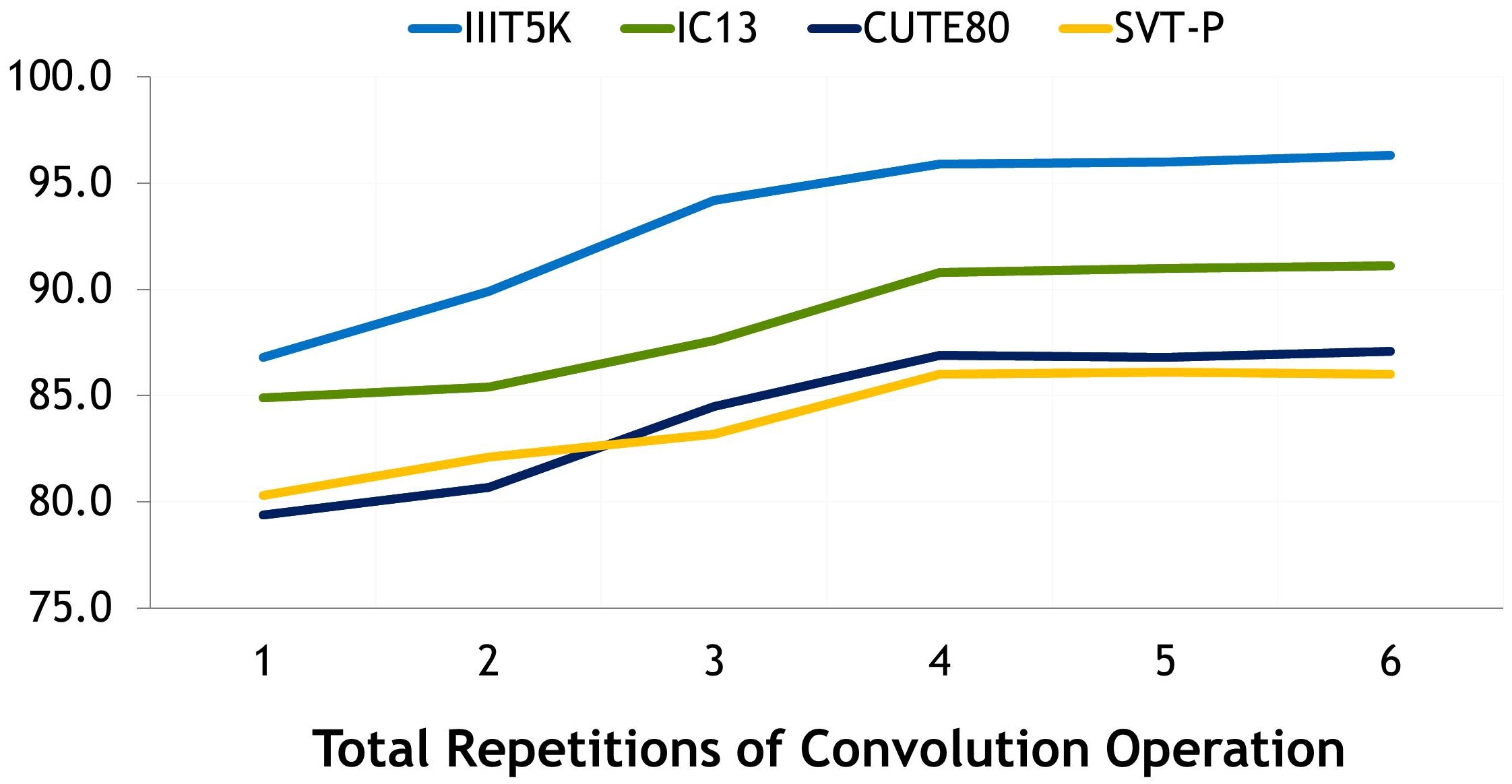}}
		\end{center}
	\end{minipage}
	\vspace{-6mm}
	\caption{\footnotesize{VAB convolution quantity analysis graph. The graph indicates that repeating the convolution operation four times in the VAB block before the segmentation-map generation yields the optimal accuracy as tested on four different datasets.}}
    \vspace{-3mm}
    \label{fig:time_analysis}
\end{figure}

\setlength{\tabcolsep}{2.0pt}
\begin{table*}[t]\small

 \vspace{-5mm}
	\caption{\footnotesize Robustness Analysis. This study demonstrates the proposed model has better robustness towards different changes in the input images. (acc: accuracy, diff.: Accuracy Difference of performance from the original dataset, change (\%): Percentage change (decrease) in the accuracy).}

	\begin{center}
	\begin{tabular}{|c|c|c|c|c|c|c|c|c|c|c|c|c|c|c|}
    \hline

& IIIT & \multicolumn{3}{c|}{IIIT-padded} & \multicolumn{3}{c|}{IIIT-r-padded} & IC13 & \multicolumn{3}{c|}{IC13-expanded} & \multicolumn{3}{c|}{IC13-r-expanded} \\ \hline
Method & acc & acc  & diff. & change (\%) & acc  & diff. & change (\%) & acc & acc  & diff. & change (\%) & acc  & diff. &  change (\%) \\ \hline

CA-FCN \cite{liao2019scene} &  92.0 & 89.3  & -2.7 & 2.9   & 87.6 & \textbf{-4.4} & 4.8 & 91.4 & 87.2 & -3.7 & 4.1 & 83.8 & -6.9 & 7.6 \\ 
DAN-1D \cite{wang2020decoupled} & 93.3 & 91.5 & -1.8  & 1.9   & 88.2 & -5.1 & 5.4 & 94.2 &  91.2 & \textbf{-3.0} & \textbf{3.2} & 86.9 & -7.3 & 7.7 \\ 
DAN-2D \cite{wang2020decoupled}  & 94.3 & 92.1 & -2.2 & 2.3   & 89.1 & -5.2 & 5.5 & 93.9 & 90.4 & -3.5 & 3.7 & 86.9 & -7.0 & 7.5 \\ 

\textbf{Ours} & \textbf{95.9} & \textbf{94.0} & \textbf{-1.9} & \textbf{2.0}   & \textbf{91.4} & -4.5 & \textbf{4.7} & \textbf{96.3} & \textbf{92.5} & -3.8 & 3.9 & \textbf{89.5} & \textbf{-6.8} & \textbf{7.1}  \\ \hline

	\end{tabular}
	\end{center}
	   
	\label{table:robustness_results}
    \vspace{-3mm}
\end{table*}

\noindent \textbf{Ablation Study.} 
We perform five different ablation experiments to analyze different components of our network.

1) \textit{Effect of VAB Block}: The VAB block provides the most important  visual attention mechanism that improves the network performance. As shown in Table \ref{table:abl_results}, the network under-performs on both regular and irregular scene-text datasets without using the VAB block. Thus, it's imperative to include the VAB block.

2) \textit{Number of Residual Units}: The number of residual units (RU) in the RS block plays an important role in better feature extraction. We experimented with different RU units quantity per RS block as shown in Table \ref{table:abl_results}. As per the results, we found five RU units per RS block to be the most effective choice with the highest accuracy.

3) \textit{Effect of S2 and S3 scale-modules inclusion}: As given in Table \ref{table:abl_results}, using the S2 and S3 scale-modules in addition to S1 increases the network effectiveness. However, adding another scale-module S4 does not enhance the accuracy significantly. Thus, the (S1, S2, S3) combination has been employed.

4) \textit{MaxLength Value Selection}: The MaxLength value has to be selected so that it covers the maximum length an output word can possibly have in a dataset. Beyond that, increasing it should not have any noticeable effect on the network efficacy. As given in Table \ref{table:abl_results}, increasing the MaxLength value from the default value of 25 does not alter the performance by much.

5) \textit{Total Convolution Operations in VAB Block}: We investigate the effect of a total number of convolution operations before the segmentation map creation. To analyze the effect, we perform convolution operations quantity experiments on four datasets (IIIT5k, IC13, CUTE80, SVT-P). The results are shown in Fig. \ref{fig:time_analysis}, where repeating four convolution operations before the segmentation map generation in the VAB block proves to be the best choice.  

\noindent \textbf{Robustness Analysis.} Here, we check for robustness of the proposed scheme against different modifications on the input images. We compare our scheme with two recent SOTA methods (DAN \cite{wang2020decoupled} and CA-FCN \cite{liao2019scene}) on two datasets (IIIT-5K \cite{mishra2012scene} and IC13 \cite{karatzas2013icdar}). Following the practice as given in \cite{wang2020decoupled}, variations introduced into these datasets are as follows:

\textbf{IIIT-padded:} 100\% padding of the input images in IIIT-5k in both horizontal and vertical direction via border pixels replication. \textbf{IIIT-r-padded:} Stretching the image vertices using a random scale value up to 20\% for both height and width respectively. Next, repetitive border pixels have been used for filling it. Finally, we crop the axis-aligned rectangles. \textbf{IC13-expansion:} The input images in IC13 are expanded into image frames with relatively extra 10\% height and width followed by cropping. \textbf{IIIT-r-expansion:} Expansion of the IC13 images using a random scale up to 20\% height and width, followed by cropping the axis-aligned rectangular images.

As shown in Table \ref{table:robustness_results}, it can be observed that the proposed method appears as the most stable and resilient to these input distortions and variations in majority cases, hence, demonstrating the robustness of our scheme.

\subsection{Qualitative Analysis}
\begin{figure}
	\begin{minipage}[b][][b]{0.32\columnwidth}
		\begin{center}
			\centerline{\includegraphics[width=1\columnwidth]{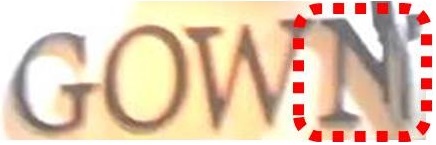}}
			\centerline{\footnotesize{GOW\textcolor{red}{IV}}}
			\centerline{\footnotesize{GOWN}}
		\end{center}
	\end{minipage}
		\begin{minipage}[b][][b]{0.32\columnwidth}
		\begin{center}
			\centerline{\includegraphics[width=1\columnwidth]{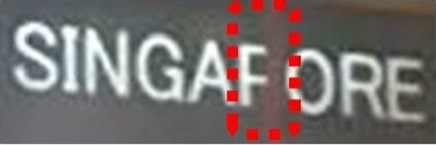}}
			\centerline{\footnotesize{SINGA\textcolor{red}{F}ORE}}
			\centerline{\footnotesize{SINGAPORE}}
		\end{center}
	\end{minipage}
	\begin{minipage}[b][][b]{0.32\columnwidth}
		\begin{center}
			\centerline{\includegraphics[width=1\columnwidth]{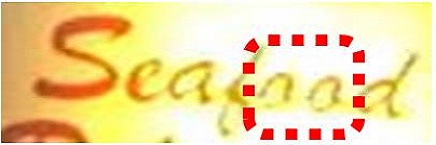}}
			\centerline{\footnotesize{Seaf\textcolor{red}{aa}d}}
			\centerline{\footnotesize{Seafood}}
		\end{center}
	\end{minipage}
	\begin{minipage}[b][][b]{0.32\columnwidth}
		\begin{center}
			\centerline{\includegraphics[width=1\columnwidth]{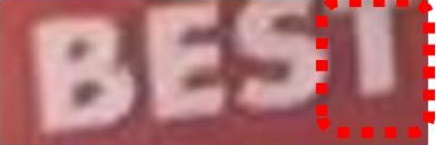}}
			\centerline{\footnotesize{BES\textcolor{red}{I}}}
			\centerline{\footnotesize{BEST}}
		\end{center}
	\end{minipage}			
		\begin{minipage}[b][][b]{0.32\columnwidth}
		\begin{center}
			\centerline{\includegraphics[width=1\columnwidth]{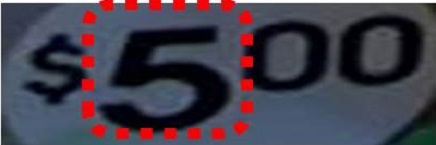}}
			\centerline{\footnotesize{\$\textcolor{red}{S}00}}
			\centerline{\footnotesize{\$500}}
		\end{center}
	\end{minipage}
	\begin{minipage}[b][][b]{0.32\columnwidth}
		\begin{center}
			\centerline{\includegraphics[width=1\columnwidth]{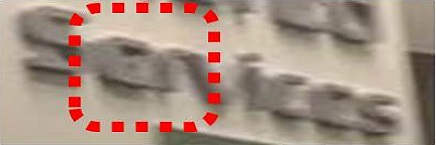}}
			\centerline{\footnotesize{S\textcolor{red}{cn}vices}}
			\centerline{\footnotesize{Services}}
		\end{center}
	\end{minipage}

		\begin{minipage}[b][][b]{0.32\columnwidth}
		\begin{center}
			\centerline{\includegraphics[width=1\columnwidth]{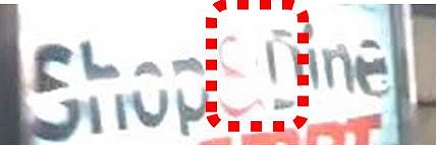}}
			\centerline{\footnotesize{Shop\textcolor{red}{8P}ine}}
			\centerline{\footnotesize{Shop\textcolor{red}{8}Dine}}
			\centerline{\footnotesize{GT: Shop\&Dine}}
		\end{center}
	\end{minipage}
		\begin{minipage}[b][][b]{0.32\columnwidth}
		\begin{center}
			\centerline{\includegraphics[width=1\columnwidth]{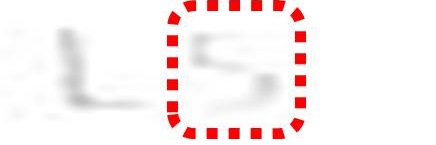}}
			\centerline{\footnotesize{L\textcolor{red}{S}}}
			\centerline{\footnotesize{L\textcolor{red}{S}}}
			\centerline{\footnotesize{GT: L5}}
		\end{center}
	\end{minipage}
		\begin{minipage}[b][][b]{0.32\columnwidth}
		\begin{center}
			\centerline{\includegraphics[width=1\columnwidth]{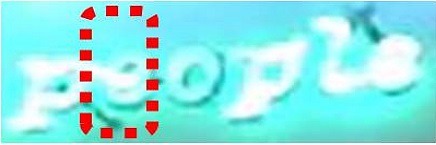}}
			\centerline{\footnotesize{p\textcolor{red}{8}ople}}
			\centerline{\footnotesize{p\textcolor{red}{s}ople}}
			\centerline{\footnotesize{GT: people}}
		\end{center}
	\end{minipage}
					
    \vspace{-1mm}
	\caption{\footnotesize{Ground truth (GT) scene-text based qualitative comparison. The first two rows demonstrate the good prediction results followed by the bad recognition cases in the last row. Under each image, the first line indicates our network text prediction without using the VAB block, whereas the second line shows our model with the VAB. The red-colored characters indicate the wrong predictions.
	}}
	\label{fig:qualityResults}
    \vspace{-4mm}
\end{figure}

Here, we present some good and bad qualitative results. We evaluate the proposed scheme with and without the visual attention block (VAB). The results are shown in Fig. \ref{fig:qualityResults}, where the first two rows indicate the good results followed by the failure cases in the last row. Following the practice in \cite{yu2020towards}, under each image, the first line shows the text recognition made by the proposed scheme without using the VAB block followed by our network text prediction with the VAB module in the second line. Characters colored as red indicate wrong predictions. As shown in the good results, the proposed scheme without the VAB block lacks the visual attention and struggles to differentiate between highly similar characters (e.g. 'e' and 'c' or 'o' and 'a') when they lack clear visual exposure, skewed perspective, or partial occlusion. The VAB block coupled with the multi-scale fusion helps in overcoming these issues and produces accurate results as shown. 

The bad results, as shown in the last row of Fig. \ref{fig:qualityResults}, mainly occur when the visual attention does not align the characters perfectly and results in failure as compared to the ground-truth (GT) recognition text.

\section{Conclusion and Future Work}
In this paper, we proposed a new multi-scale and scale-wise visually attended text recognition network to address key scene-text challenges. The multi-scale feature extraction and visual attention have been performed in parallel to utilize different feature scales explicitly in a more effective way. The network also undergoes multi-scale fusion with each other to develop the coordinated information. Experimental evaluation on standard benchmarks indicates better accuracy in most cases as compared to the SOTA methods.

One of the key limitations of the proposed network is that it is using simpler inter-scale fusion. In the future, we aim to investigate more sophisticated fusion techniques. Moreover, our current work focuses only on the offline recognizer design, we will investigate the efficiency and computational cost aspects for real-time and real-world applications in the future.

\section{Acknowledgements}
This work was done while Usman was interning at SIE Global R\&D. T. Kim was supported in part by the National Science Foundation (NSF) under grants CNS1955561 and AST2037864. G. Wang is supported by the Natural Sciences and Engineering Research Council of Canada (NSERC) under grant RGPIN-2021-04244.

\balance
\bibliographystyle{IEEEtranS}
\bibliography{egbib}
\balance

\end{document}